%% file: templateArxiv.tex
\documentclass{article}

\usepackage{amsmath,amssymb,amsfonts}
\usepackage{algorithmic}
\usepackage{graphicx}
\usepackage{textcomp}
\usepackage{subfigure}

\usepackage{PRIMEarxiv}

\usepackage[utf8]{inputenc} 
\usepackage[T1]{fontenc}    
\usepackage{hyperref}       
\usepackage{url}            
\usepackage{booktabs}       
\usepackage{amsfonts}       
\usepackage{nicefrac}       
\usepackage{microtype}      
\usepackage{lipsum}
\usepackage{fancyhdr}       
\usepackage{graphicx}       
\graphicspath{{media/}}     

\title{OrthoGAN:High-Precision Image Generation for Teeth Orthodontic Visualization}

\author{
  Shen Feihong \\
  Jilin University \\
  gregoryfeihong@gmail.com \\
  \And
  Liu jingjing\\
  Zhejiang University \\
  12221107@zju.edu.cn \\
  \And
  Lou jianwen\\
  Zhejiang University \\
  jianwen.lou@zju.edu.cn \\
  \And
  Li Haizhen \\
  Shanghai Jiao Tong University School of Medicine \\
  569286671@qq.com \\
    \And
  Bing Fang \\
  Shanghai Jiao Tong University School of Medicine \\
  fangbing@sjtu.edu \\
    \And
  Ma Chenglong \\
  Zhejiang University \\
  mchenglong@chohotech.com
    \And
  Jin Hao \\
  Harvard University \\
  jinhao@g.harvard.edu
    \And
  Yang Feng \\
  Angelalign Research Institute \\
  fengyang@angelalign.com \\
  \And
  Zheng Youyi \\
  Zhejiang University \\
  youyizheng@zju.edu.cn
}

\begin{document}
\maketitle

\begin{abstract}
    Aligning the teeth of a face image with a specified teeth shape is crucial in orthodontic treatment visualization. This task falls within the realm of conditional image editing, an area that has witnessed numerous methodological advancements in recent years. Despite this progress, existing methods face challenges in preserving the identity of the original teeth, particularly influenced by factors such as 3D teeth shape and lighting conditions. To tackle this issue, we introduce a novel style-based generative model designed to synthesize identity-preserving teeth images. Our model takes a user's frontal smile face image and their 3D scanned teeth model as input. It then transforms the 3D model into multi-modal conditions, including a binary mask, contour image, and depth map. The model outputs a face image with teeth that match the given 3D scan while aligning with the original image's contextual identity. The proposed method surpasses state-of-the-art teeth alignment techniques, exhibiting improved visual quality. Moreover, a rigorous user study, involving both doctors and patients, provides compelling evidence of the superior performance of the proposed method.
\end{abstract}

\keywords{Deep Learning, Computer simulation, teeth alignment}

\section{Introduction}
\label{sec:introduction}
\input{figure/fig1/figure1}

Teeth alignment of the frontal smile face is getting more and more attention \cite{yang2020iorthopredictor, chen2022orthoaligner}. Patients who choose invisible teeth alignment have the willingness to know the effects of teeth alignment in different treatment stages. The natural goal of this task is to replace the teeth region of the smile face images with the generated teeth images. Different from other inpainting tasks \cite{wang2022dual}, the generated teeth images in this task should be strictly aligned with the patient's actual 3D teeth model and have the natural outlooks simultaneously. Due to the complicated data formation in this task, the previous works \cite{yang2020iorthopredictor} also fail to efficiently generate an aligned frontal smile image with high quality. Therefore, an accurate and efficient framework that can automatically predict high-quality smile simulation in photographs is needed for orthodontic visualization.

 Orthodontic visualization mainly includes two sub-tasks: the pose fitting of scanned 3D teeth models and the generation of face images in the region of interest. The solutions of these two sub-tasks in previous work \cite{yang2020iorthopredictor} contain irrational structures, thereby leading to relatively low accuracy and suboptimal outcomes. In this paper, we introduce novel approaches to address these issues. Specifically, we leverage differentiable rendering on the pose fitting stage and transfer the aligned 3D data to other modalities, including 2D teeth silhouette, depth image, and label mask. Our framework efficiently makes good use of these modalities to generate high-quality teeth images with our synthesis component. It should be noted that, in this paper, we do not include the content of 3d teeth auto-alignment.

The most essential part of alignment visualization is partial face image manipulation.  Diffusion models \cite{dhariwal2021diffusion, karras2022elucidating,ho2022video} have demonstrated the ability to generate high-fidelity human face images. While several works propose some methods to reduce the inference speed and add controllable generating conditions, like ControlNet \cite{zhang2023adding}, most of these models require a large number of iterations to generate face images and a higher order of magnitude of training resources compared to style-based generation models. In figure \ref{Fig1}, we test some public image-to-image models, and the results are not suitable for clinical applications. Similar to previous work\cite{yang2020iorthopredictor}, we introduce an adversarial mechanism and novel training method to our synthesis model $OrthoGAN$. Instead of the diffusion model, in the last decade, generative adversarial networks (GAN) have been utilized in many real-world applications \cite{mirza2014conditional, brock2018large} and have achieved state-of-the-art performances. Many architectures of the GAN have been proposed to improve the fidelity and diversity of the generated signal including WGAN \cite{arjovsky2017wasserstein}, BigGAN \cite{brock2018large} et al. In recent years, StyleGAN \cite{karras2019style} has been known for its ability to generate realistic images in various domains. EditGAN \cite{ling2021editgan} applied it to generate realistic facial images, which allow users to edit images by modifying highly detailed part segmentation masks. Inspired by the success of multi-modal GAN \cite{Huang2021MultimodalCI}, we propose our $OrthoGAN$ network based on the style generation method, which combines the teeth geometry information from the 3D model and teeth visual characteristics from the RGB image to generate the corresponding after-treatment smile face photograph. Specifically, our novel architecture in the network structure and training method enables the model to fit the slight edge in input maps and the customized style-based architecture in our network allows the model to simulate complicated illumination conditions \cite{luo2023pseudo} on the faces of the teeth surpassing the former performance.

In summary, our main contributions are:
\begin{itemize}
    \item We introduce differentiable rendering to the pose fitting of the 3D teeth model and surpass the performance of the previous method in both efficiency and accuracy. In particular, our edge-rendering-based fitting can be used on other fitting tasks that have object occlusion or lack RGBD feature information.
    \item We design a new conditional generation model to achieve teeth alignment visualization in the oral domain. It guarantees generation fidelity with a relatively small size and fast speed. This architecture can be applied to other conditional inpainting tasks that involve time series inputs.
\end{itemize}

\input{figure/fig2/figure2}  

\section{Related Works}

In this section, we review the previous research in the areas of differentiable rendering, and human face manipulation, which are highly related to our problem.

\subsection{Differentiable Rendering}
In traditional rendering, a series of parameters such as geometry, lighting, material, and camera position are passed to a rasterization or ray tracing renderer, which outputs an image after a series of pipeline operations. Due to its difficulty in reverse derivation, differentiable rendering came into being. It introduces differentiable operations in the rendering process so that it can adjust the rendering parameters through gradient backpropagation to optimize the difference between the output result and the target \cite{kato2020differentiable}.

In recent years, differentiable rendering has been used to solve various 3D-related problems in computer vision and computer graphics, especially the reconstruction of mesh\cite{jiang2022selfrecon, deng2022reconstructing} and texture based on mesh \cite{lin2022multiview}. Supervised learning from real 3D shapes is a straightforward approach in 3D object reconstruction from images, which requires annotation of the 3D shapes corresponding to the images. Some work proposes to replace 3D annotations with 2D annotations \cite{wu2017marrnet, ramakrishna2012reconstructing}, which can reduce the cost of labeling 3D objects. For example, Kanazawa uses the distance of the projection of the key points of the object to reconstruct the shape of the object \cite{yu2021pixelnerf}. Bogo recovers human pose using joint point locations predicted from 2D images \cite{hasson2020leveraging}. In addition to image key points, there are also works using 2D mask \cite{baek2019pushing} and heatmap \cite{zhang2021interacting} as supervisory signals to assist object reconstruction more accurately. Due to the practical application of human faces in the game and animation industries, many works focus on using differentiable rendering for face reconstruction \cite{yu2023unsupervised, wang2021deep, papantoniou2023relightify}. Wu et al. \cite{wu2020unsupervised} proposed a fully unsupervised method to learn to reconstruct real faces from face images by exploiting shadow information. However, different from face images, the keypoints on the teeth images are much harder to precisely detected due to the scarcity of features and occlusion between the lips and teeth.

In the application of differentiable rendering, a reasonable gradient loss function is crucial. In this paper, we design a special rendering pipeline to optimize the camera pose and internal parameters by the difference of tooth edges. 

\subsection{Human Face Manipulation}
Face editing is to change certain features of a specific face, such as changing gender \cite{shen2020interpreting}, age\cite{yao2021high}, hair color \cite{wei2022hairclip}, and so on, which can realize the changes of abstract faces and transform faces. It has a wide range of applications in face beautification, face prediction, and other fields. In the early days, researchers used methods based on statistics \cite{terada2009automatic}, based on gradient techniques \cite{yu2004mesh}, based on prototypes or physical models. However, these methods have certain limitations. Prototype-based methods \cite{rowland1995manipulating} use the average difference between different attributes for mode transfer, so that the generated faces do not have individual characteristics. Physical model-based \cite{aizawa1995model} methods perform parametric modeling on shape and texture, which usually requires a large number of training samples. With the development of data science, face editing is regarded as a regression problem. Jin \cite{jin2013eyeglasses} proposed a method to remove face glasses, training a linear regression model from face images with glasses and image samples without glasses, but its performance largely depends on the quality of the samples. 

With the prosperity of deep learning, generative models such as GAN \cite{IanGoodfellow2014GenerativeAN} and VAE \cite{IanGoodfellow2014GenerativeAN} have been proposed. Through generative models, low-dimensional data is converted to a high-dimensional image data domain, and face generation is no longer difficult \cite{dhariwal2021diffusion, karras2022elucidating,ho2022video}. At present, the most advanced face generation models, such as BigGAN \cite{brock2018large} and StyleGAN \cite{karras2019style}, can already create high-resolution realistic face images that do not exist in the real world. Among the generative models for image editing, StyleGAN \cite{karras2019style, karras2020analyzing} performs most prominently and is widely recognized as a model for generating high-quality faces. It introduces a style-based multi-scale generator to map images in a latent space. Encoders such as pSp \cite{richardson2021encoding} and e4e \cite{tov2021designing} are trained to encode target face images into latent codes for further editing. Subsequent works \cite{abdal2019image2stylegan,liu2022deepfacevideoediting} like StyleEX \cite{yang2023styleganex} added some constraints to edit image details, such as wrinkles, skin color, glasses, etc., which generate high-definition face images. However, there are few fine-grained local image generation and fusion, such as tooth and hair. Several previous works \cite{liu2022towards} suffer from the lack of hard control in other face regions.

Like UPHDR-GAN \cite{li2022uphdr}, we do not have pair images for training in this work. Compared with prior works, we focus on manipulating the mouth region of human face images and present a new generator based on StyleGAN, which can be regarded as the decoder in our encoder-decoder framework.

\section{Methodology}
Given a face photograph $\mathcal{X}$ of a patient with visible misaligned teeth, the CAD model of scanned teeth $\mathcal{T}_{0}$, the orthodontics plan the alignment treatments represented as new teeth models $\left\{\mathcal{T}_{i}\right\}_{i=1}^{N}$. Our purpose is to visualize the treatment on the teeth region in partial image $x_0$ while keeping other parts the same. Omitting detecting the position of $x_0$ in $\mathcal{X}$, This task can be split into two major parts: pose fitting of $\theta$ and conditional image inpainting of $x_i$. We denote $\theta$ as fitting parameters and $x_i$ as the inpainting result of $\mathcal{T}_{i}$ on $x_0$.

\subsection{Analysis of the Previous Limitation}
In the former work\cite{yang2020iorthopredictor}, the teeth alignment visualization was formulated as:
\begin{equation}
\mathcal{L}(x, \theta, \phi)=\mathbb{E}_{q(z \mid x)}(-\log p(x \mid z, g))+D_{k l}(q(z \mid x) \| p(z)),
\end{equation}
where $\theta$ and $\phi$ denote the the parameters of their generative network $\mathcal{N}$ and encoder $\mathcal{M}$, and $D_{k l}$ refers to the KL divergence. This formulation is based on a strong consumption that the in-mouth appearance $z$ is independent of the geometry $g$. This consumption is unreasonable because the shape of teeth influences the in-mouth appearance. Thus the simple traditional U-Net architecture without style structure in previous work \cite{yang2020iorthopredictor} fails to handle this task. To circumvent this problem, the former work extract the teeth silhouette $g_y:=\{g_u, g_l\}$ to independently represent the teeth geometry and utilize style-based convolution and separate encoder $\mathcal{N}_{enc}$ to avoid the spatial attribution in $z$. These strategies bring new problems. First, the teeth silhouette $g_y$ contains little information to represent teeth geometry. Second, the double-encoder synthesis network structure is unstable in some special cases and easy to collapse in the training stage. To figure out the first problem, we add a new input map to represent the geometry of the teeth: $g_d$, thus our $g_y$ becomes $\{g_u, g_l, g_d\}$. The previous work \cite{yang2020iorthopredictor} also mentions a similar input map in the evaluation of $TSynNet$. However, in their comparison, the output of $TSynNet$ is based on well-posed teeth silhouette while the output of the control group \cite{xie2015holistically} is based on wrong fitted teeth pose which we believe is unfair. This comparison also reveals to us that the pose-fitting algorithm in previous work is unreliable, even with their interactive interface. We introduce our new pose fitting algorithm and new input geometry map in Sec. \ref{sec:3.2}. We also propose a new conditional synthesis model to replace the $TSynNet$ in \cite{yang2020iorthopredictor}. The structure and training method is introduced in Sec. \ref{sec:3.3}.

\subsection{Teeth Pose fitting}
\label{sec:3.2}
As shown in Fig. \ref{Fig1}, in stage 2, we project the 3D teeth $\mathcal{T}_{i}$ arranged by orthodontists to the input modalities needed in stage 3. Since the face image $\mathcal{X}$ is not calibrated, we need to predict the parameters for camera extrinsic, intrinsic, and radial distortions. The optimized parameters $\theta$ in our work also contain the offset of lower teeth towards upper teeth. Different from other pose-fitting tasks, most teeth images in portrait exhibit all-white texture without enough RGB or depth information to fit. The occlusion from lips and bad intraoral lighting conditions also require the fitting method to be more robust. So we customize a differentiable depth shader to replace the EM algorithm in former work \cite{yang2020iorthopredictor}.  We render a depth image by the differentiable renderer $r$ with our customized shader. The depth image is transferred to a teeth contour image similar to $g_y$ within the region $g_m$. Since the whole render process is differentiable, we can iteratively fit the teeth position guided by the gap between $g_y$ and the rendering result. By altering the learning rate, we can first optimize the translation matrix and then update the rotation matrix, focal length, and offset guided by the l2 loss function:
\begin{equation}
L_{\theta}:=\left\|g_u+g_l-g_m*r_{\theta}\left(T_0\right)\right\|_{2}^{2}
\end{equation}
This edge-rendering-based fitting method allows us to update more parameters than previous work thus our fitting result has higher fidelity to the image $X$. Our edge-based optimization also circumvents the influence of inaccuracy detected keypoints in previous work. After getting the best rendering parameters, we can render teeth model $\mathcal{T}_{i}$ to depth image $g_d^i$ and transfer it to $g_u^i$ and $g_l^i$ with a common shader.
\subsection{OrthoGAN}
\label{sec:3.3}
\input{figure/fig3/figure3}  

$OrthoGAN$ is a conditional synthesis network trained by adversarial, perceptual, and control loss. We apply $OrthoGAN$ to generate natural images of the human face in the mouth region. To extract the geometry and visual feature from input data, our $OrthoGAN$ contains an encoder $\mathcal{E}$ which projects different input data into the latent space of decoder $\mathcal{D}$.

\textbf{Input data.} The input of $OrthoGAN$ can be regarded as a 7-channel image, which is divided into a 4-channel image and a 3-channel image in the encoder $\mathcal{E}$. The first 4-channel image is $input_{geo}:=((1-g_m)*x_0+g_u+g_l+g_d)\oplus g_m$ concat with $g_m$ and the 3-channel image is $input_{style}:=g_t*x_0$ where $g_t$ and $g_m$ denotes the segmentation result of teeth and mouth cavity. $g_u$ and $g_l$ represent the edge silhouette of upper and lower teeth respectively.  

\textbf{Encoder $\mathcal{E}$.} As we mentioned before, the encoder $\mathcal{E}$ takes $input_{geo}$ and $input_{style}$ to generate the input features of decoder $\mathcal{D}$. When $\mathcal{E}$ gets $input_{geo}$, the network applies an optional convolution layer to unify the channel number of $input_{geo}$. After that, the extraction module of $\mathcal{E}$ accepts the unified image and appends the middle-generated feature in an empty list. To circumvent the influence of $(1-gm)*x_0$ in $input_{geo}$, we apply convolutional layers on every element in this feature list. The last element will replace our decoder's constant input, which is built upon StyleGAN \cite{karras2019style}. Other elements in the feature list will be added on the corresponding middle feature map before processing by the StyleConv layer in our decoder $\mathcal{D}$. 

When $\mathcal{E}$ meets $input_{style}$, the image is directly cast into the extraction module without the conditional convolutional layer. Different from the elements in the feature list of $input_{geo}$, the elements in the feature list of $input_{style}$ apply the average pooling and full connection layer after the convolution to minimize the geometry information in the feature map. The output feature map list can be regarded as the latent codes of the decoder $\mathcal{D}$ to keep the style of the generated teeth image aligned with the teeth in $x_0$.

\textbf{Decoder $\mathcal{D}$.} The decoder $\mathcal{D}$ was built upon the structure of the generator of StyleGAN but it takes a $16\times16$ latent space instead of a constant input. As shown in Figure ref, the latent space is the feature result extracted from the $input_{geo}$ without the information from $input_{style}$. In the following blocks, the feature maps generated by $input_{geo}$ are added to the middle convolution results. The latent codes from $input_{style}$ perform convolution with added maps after passing the Mod\&Demod modules. Similar to the StyleGAN, the output of the decoder $\mathcal{D}$ is the summarization of results from to-RGB layers.

\textbf{Postprocess.} In both the training and testing stages, before being cast into the discriminator, the final result of OrthoGAN replaces the generated lip and skin region with the region in the original image $x_0$. Then the discriminator can detect the potential inconsistency in the final result and force the OrthoGAN to learn to blend naturally in the junction of lips and teeth. 

\textbf{Loss function.} We train the encoder $\mathcal{E}$ and the decoder $\mathcal{D}$ in different stage. The training of the decoder $\mathcal{D}$ follows StyleGAN with the adversarial loss. The OrthoGAN loads the pre-trained decoder's weight and trains the encoder with the loss:

\begin{subequations}
\begin{align}
&\mathcal{L}_{\mathrm{Ortho}}=\mathcal{L}_\text{adv}+\lambda_{1}\mathcal{L}_\text{rec}+\lambda_{2} \mathcal{L}_\text{reg} \\
&\mathcal{L}_{rec}=LPIPS\left(x_{0}, \hat{x}_{0}\right)+\lambda_{3}\left\|x_{0}-\hat{x}_{0}\right\|_2^2
\end{align}
\end{subequations}
where $\mathcal{L}_\text{adv}$ is an adversarial loss to improve the realism. $\mathcal{L}_\text{rec}$ is the reconstruction loss which contains perceptual loss and pixel loss and encourages the generated teeth to have the same appearance as the original teeth. $\mathcal{L}_\text{reg}$ \cite{richardson2021encoding} evaluates the distance between the latent code predicted by the encoder and the average latent code. 

\section{Experimental Results}

\textbf{Datasets.}
We get two modal data from our collaborated medical institutions: a face photograph of a patient with visible misaligned teeth $\mathcal{X}$ and a list of 3D teeth objects $\left\{T_i\right\}_{i=1}^k$ representing the changing of the patient's teeth during the alignment. We evaluate our method on the dataset processed from these data. The dataset contains 2000 cases, covering a wide range of patients with different ages, genders, and teeth appearance. We use 1900 of them to train our network and select 100 cases as the test set.

\noindent\textbf{Implementation details.}
We set $\lambda_{1}=0.01$, $\lambda_{2}=1$ and $\lambda_{3}=0.1$ in our loss functions. All networks use the Adam optimizer with learning rates = 0.0001. The decoder takes 25000 training steps and the encoder of OrthoGAN and other comparison networks takes 15000 training steps. It empirically takes 200 steps for the differentiable rendering method to fit the teeth pose. The mouth region image is set to $256 \times 256$ resolution in our experiments. All experiments run on a server with eight Nvidia 3090 GPUs, an AMD EPYC 7402 24-core processor, and 252 GB RAM. 

\noindent\textbf{Baselines.}
We evaluate our method on two aspects: (1) \textbf{Pose fitting}: we run the fitting algorithm in iOrthoPredictor \cite{yang2020iorthopredictor} to show the robustness of our method. (2) \textbf{Image inpainting}: we compare the mouth image inpainting result between StyleEX \cite{yang2023styleganex}, TSynNet in iOrthoPredictor, and our OrthoGAN. As we mentioned before, StyleEX is a simple and effective method for characterizing unaligned faces. We apply it to the inpainting task with the mask-to-face translation setting combined with the sketch-to-face translation setting. We exhibit the effectiveness of our design through ablation studies. Our pipeline can handle challenging cases while the former method fails to generate correct results. 

\noindent\textbf{Fitting result}
We randomly selected 200 patients with different ages, genders, and smile conditions as our test group. We regard 10 of them as challenging cases because of the large occlusion of lips and teeth and the offset of upper and lower teeth. We perform the point-based EM algorithm and our differentiable pose fitting on the test set. According to our statistics, the EM algorithm failed 50 cases, and 20 of them were in the challenging set. Our method failed 3 cases, and all of them were in the challenging set.

\noindent\textbf{Inpainting result}
   As shown in figure \ref{Fig6}, some results of TSynNet have black spots on the tooth and lack natural depth, especially in the corner of the mouth. These two flaws reveal that the single silhouette of teeth fails to represent the full geometric information needed by the generation network. To prove this assumption, we remove the depth $g_d$ in $input_{geo}$ and our OrthoGAN also shows similar black spots and teeth with unnatural depth in the corner. 

Our result of OrthoGAN also shows better blending results and closer teeth color than the other two models. The better blending results contribute to the discriminator of our adversarial loss which forces the style of generated teeth to align with the lip and skin region concatenated in the post-process. In the meantime, our ablation study proves that with the same length latent code, the replacement of lip and skin regions in the post-process also helps the decoder $\mathcal{D}$ reconstruct the color of teeth. In contrast, the inpainting task without the completed input information on lip and skin seems like an ill-posed problem in the former work \cite{yang2020iorthopredictor}.

Given the original teeth image $x$ as target and the $input_{geo}$ from the initial teeth arrangement $\mathcal{T}_{0}$, we can quantitively evaluate the teeth visual reconstruction ability of different generation models: $\mathcal{L}_{visual}:=MSE\left(g_m*x,g_m*\hat{x}_{0}\right)$. With a well-trained teeth edge detection network $\mathcal{D}$, we can statistic the edge accuracy in the generated teeth images: $\mathcal{L}_{edge}:=MSE\left(g_l^i+g_d^i,\mathcal{D}\left(\hat{x}_{i}\right)\right)$. The $MSE$ denotes the mean squared error and $\hat{x}$ is the generated images. We list the results in Table \ref{tab:3}. The $DR$ denotes that the inputs of $TSynNet$ are based on the fitting results by differentiable rendering.

\input{figure/fig4/fig4}
\input{figure/fig5/new_fig5}
\input{table/table3}

\noindent\textbf{User study} 
To evaluate the generation quality of the three models on teeth image inpainting, five doctors and five patients were collected to validate the simulation results of the three models. The images were evaluated in three aspects including simulation authenticity, simulation aesthetics, and image quality. The results were evaluated in a blind method by five doctors and five patients. The score ranges from 0 to 10 from low to high. To avoid subjective bias among different doctors and patients, we regard the average score of several patients as the final evaluation result. 

The final data were shown as means (with standard deviations). Groups were compared by one-way analysis of variance, and the significance of mean difference within (intra) and between (inter) groups was performed using the PSD post hoc test after ascertaining normality by the Shapiro-Wilk test and the homogeneity of variance between groups by the Levene test. A two-tailed $P$ value less than 0.05 $(P < 0.05)$ was considered statistically significant. All analyses were performed using SPSS software (Windows version 28.0; SPSS Inc., Chicago, IllIL, USA).

The evaluation results of the doctors are shown in Table \ref{tab:1}. The scores of simulation authenticity and image quality in OrthoGAN are statistically higher than those of TSynNet and StyleEX. The simulation aesthetics of OrthoGAN and StyleEX show almost no difference but are statistically higher than that of the TSynNet. In conclusion, the OrthoGAN shows the best performance in the view of orthodontics. 

The evaluation results of the patients are shown in Table \ref{tab:2}. OrthoGAN and TSynNet show almost no difference in simulation authenticity. The scores of OrthoGAN and TSynNet in image quality were higher than StyleEX, which indicates that StyleEX shows the worst performance in the eyes of doctors and patients. 
\input{figure/fig6/figure6}

\input{table/table1}
\input{table/table2}

\section{Conclusion}
In this study, an image was generated based on the before-treatment frontal face image and the after-orthodontics 3D teeth model. We transfer the input modalities for the quality and efficiency of image generation. In the modal transfer stage, our differentiable rendering is more efficient and precise.  Our method does not rely on any preconditions and its optimized parameters are explainable. The gradient descent which has been proven to be one of the most powerful optimizing methods \cite{ruder2016overview} accelerates our rendering process and makes it a lot faster than the EM algorithm in previous work.

The most important contribution of this work is the design of our generation model: OrthoGAN. We take the single encoder architecture to form a UNet structure and stabilize the training of our network. The training of TSynNet which contains double encoders fails many times in our experiment. Previous work \cite{yang2020iorthopredictor} proves that the large size of latent space helps the reconstruction of teeth geometry in the final result. However, due to the leak of visual information in the input data, they only applied the $4 \times 4$ latent space while we solved this problem and our latent space is $16 \times 16$. The larger latent space brings more precise teeth contour in our experiment as shown in table \ref{tab:3}. The post-process module in our OrthoGAN also helps to generate smile images with more realistic teeth including color, lighting, and perspective effects. The concatenate method forces the network to blend the lips and teeth and the latent code to focus on the appearance of the original teeth. This method can be applied to other style-based inpainting tasks. Apart from the network structure, the selection of input modalities also contributes to the exhibition. Many layers in OrthoGAN are very similar to the style-based layers in StyleEX. However, like the TSynNet, the entangled and confused input modalities lead to some obvious defects in the generated teeth image. While we should provide plenty of control information to the networks, the influence between different modalities and differences in the training and testing stages should be taken into consideration in multi-modal conditional generation tasks.

There are still some limitations to our approach. Our method only focuses on the mouth region, thus the facial growth that could alter during the orthodontic treatment is untouched. Simulation-based methods \cite{RolfMKoch1996SimulatingFS} may be exploited to solve this issue by considering the change of facial bones during the treatment. 

On this basis, we conclude that this study proposed a high-precision visualization framework for orthodontic smile simulation. The system uses a series of image-based facial editing techniques to generate the most natural and accurate simulation results for users to date. Experiments and perceptual user studies in different settings show the effectiveness of our approach in predicting the treatment effect in digital orthodontics.

With the rapid development of image generation, our work focuses on a feasible solution to a specific problem of orthodontic alignment. It convincingly reveals that deep learning and artificial intelligence possess great potential to develop more intelligent and efficient digital dental solutions. This may be considered a promising aspect of digital orthodontics and facial image manipulation. 


\section*{Acknowledgment}
We strongly acknowledge the invaluable support of doctors and patients. The authors have no competing interests to declare that are relevant to the content of this article.





\end{document}

%% file: figure/fig1/figure1.tex
\begin{figure}[htbp]
\includegraphics[width=0.5\textwidth]{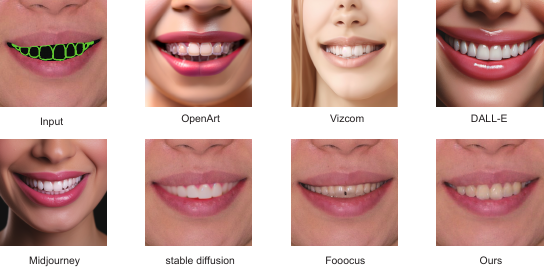}
\centering
\caption{The result from popular diffusion-based text-to-image softwares. We set the same prompt and provide the same input if the software contains the inpainting or img2img function. }
\label{Fig1}
\end{figure}

%% file: figure/fig2/figure2.tex
\begin{figure*}[htbp]     
\centering                      
\includegraphics[width=\textwidth]{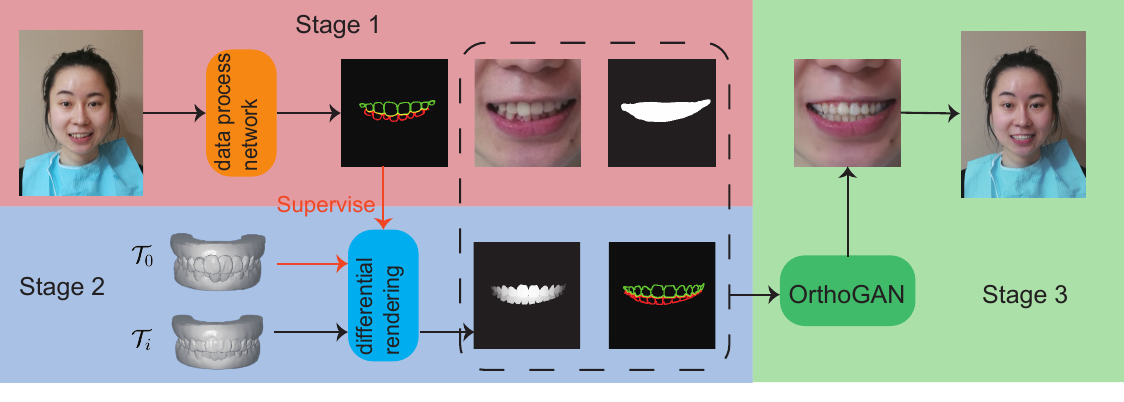}  
\caption{\textbf{Overview of our framework.} $T_0$ is the initial scanned model of patients' teeth. The parameters used to render the following $T_i$ derive from the result of the differential rendering of $T_0$. The orange arrows denote the optimization of camera parameters in the fitting stage.}      
\label{Fig2}                     
\end{figure*} 

%% file: figure/fig3/figure3.tex
\begin{figure*}[htbp]
\includegraphics[width=\textwidth]{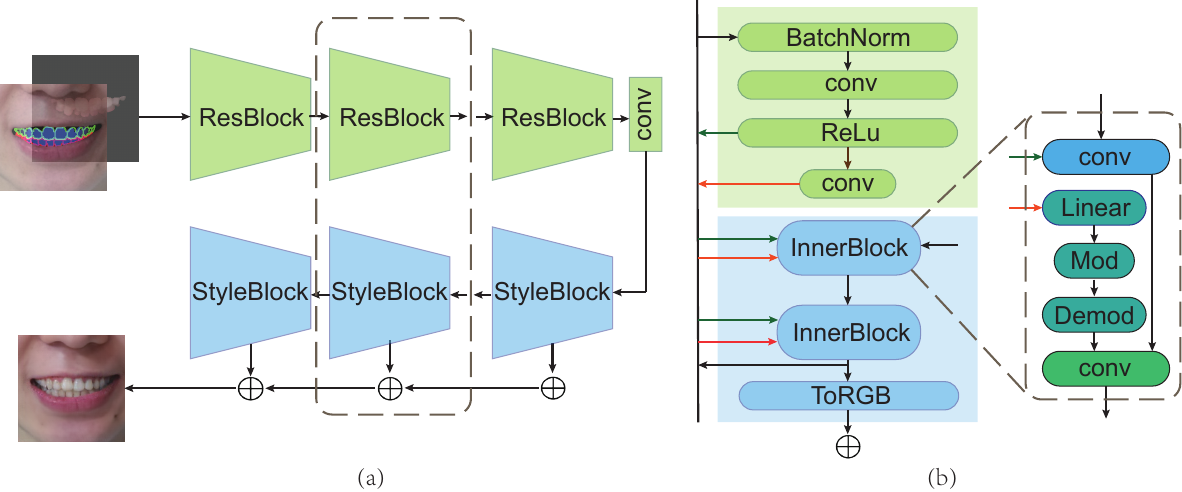}
\caption{ \textbf{The architecture of our OrthoGAN.} (a) The StyleBlock is based on the style generation structure with modulate and demodulate layers. The ResBlock and part of the StyleBlock constitute a UNet-liked structure. All ResBlock formed the encoder $\mathcal{E}$ and all StyleBlock formed the decoder $\mathcal{D}$. (b) The inner structure of ResBlock and StyleBlock. The middle arrows denote the transfer of the feature maps between two corresponding blocks. For simplicity, learned weights and noises of style mechanism are omitted.}
\label{Fig3}
\end{figure*}

%% file: figure/fig4/fig4.tex
\begin{figure*}[htbp]
\includegraphics[width=\textwidth]{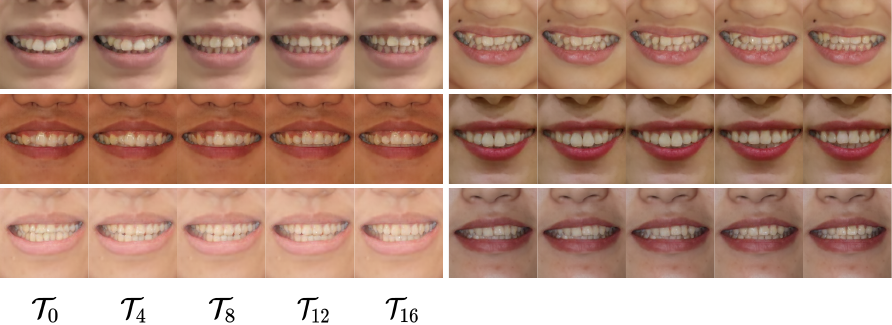}
\centering
\caption{Dynamic inpainting results of OrthoGAN. We selected several treatment alignment visualizations with steps from $\mathcal{T}_\text{0}$ to $\mathcal{T}_\text{16}$, and all the alignment simulation results can form a video to illustrate the movement of the teeth.}
\label{Fig4}
\end{figure*}

%% file: figure/fig5/new_fig5.tex
\begin{figure*}[htbp]
\includegraphics[width=\textwidth]{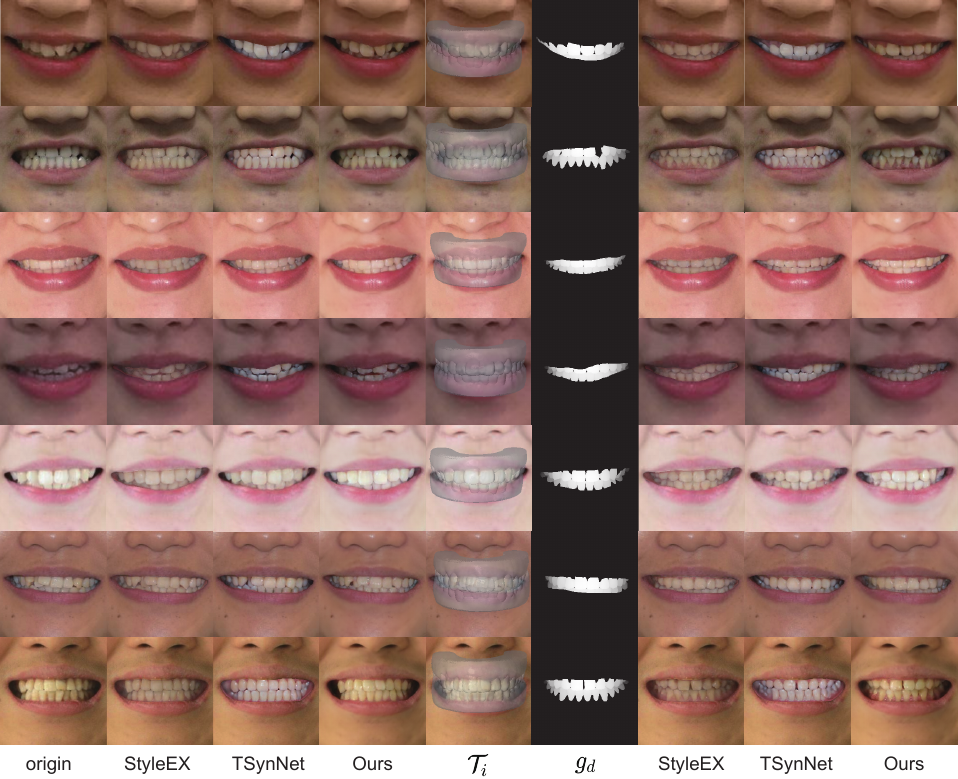}
\centering
\caption{Comparison on origin teeth reconstruction and new alignment visualization. The first column is the original teeth image. The second, third, and fourth column is the reconstruction result from StyleEX, TSynNet, and our OrthoGAN. The fifth column is the $\mathcal{T}_{i}$ and the sixth column is the rendered depth image of corresponding teeth. The seventh, eighth, and ninth column is the alignment result from StyleEX, TSynNet, and our OrthoGAN.}
\label{Fig5}
\end{figure*}

%% file: table/table3.tex
\begin{table}[!t]
\caption{Comparison of quantitative results from different frameworks on the test patient set.}
\begin{tabular}{@{}l|llll@{}}

& StyleEX  & TSyn(DR)  & Ortho w/o post  & Ortho \\ \midrule

$\mathcal{L}_{edge}$ & 0.26  & 0.24 & 0.22 & \textbf{0.21}  \\
$\mathcal{L}_{visual}$ & 1.01 & 1.01 & 1.00 & \textbf{0.86}

\end{tabular}
\label{tab:3}
\end{table}

%% file: figure/fig6/figure6.tex
\begin{figure}[htbp]
\includegraphics[width=0.5\textwidth]{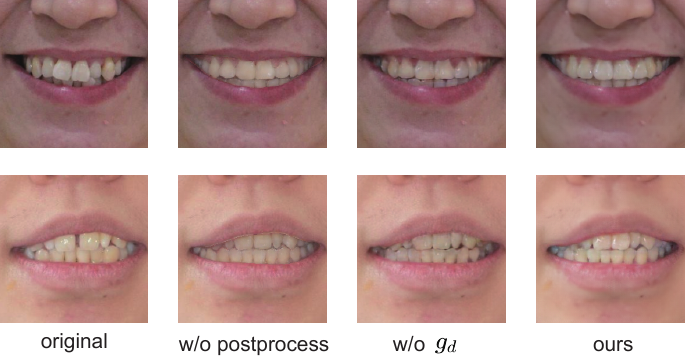}
\caption{ \textbf{Ablation experiments}. We show the generation results from the models without post-process and $g_d$ in the $input_{geo}$. }
\label{Fig6}
\end{figure}

%% file: table/table1.tex
\begin{table}[!t]
\caption{Scores statistics to the simulation results rated by orthodontics.}
\begin{tabular}{@{}l|llll@{}}
                        & OrthoGAN  & TSynNet   & StyleEX       & P    \\ \midrule
Simul-authenticity & \textbf{9.05±0.64} & 8.93±0.36 & 7.12±0.45 & 0.001         \\
Simul-aesthetics   & 7.97±0.65 & 6.83±0.38 & \textbf{8.06±0.58} & 0.0002   \\
Image quality      & \textbf{7.99±0.34}  & 7.74±0.38 & 6.88±0.38 & 0.0005     
\end{tabular}
\label{tab:1}
\end{table}

%% file: table/table2.tex
\begin{table}[!t]
\caption{Scores statistics to the simulation results rated by patients.}
\begin{tabular}{@{}l|llll@{}}
                        & OrthoGAN  & TSynNet   & StyleEX       & $P$      \\ \midrule

Simul-authenticity & \textbf{8.03±0.36} & 7.95±0.37 & 6.91±0.45 & 0.004     \\
Simul-aesthetics   & 7.71±0.44 & 6.90±0.40  & \textbf{7.88±0.38} & 0.003     \\
Image quality      & \textbf{7.79±0.42} & 7.71±0.40 & 6.96±0.36 & 0.0002     
\end{tabular}
\label{tab:2}
\end{table}